\documentclass{article}

\usepackage{arxiv}

\usepackage[utf8]{inputenc} 
\usepackage[T1]{fontenc}    
\usepackage{hyperref}       
\usepackage{url}            
\usepackage{booktabs}       
\usepackage{amsfonts}       
\usepackage{nicefrac}       
\usepackage{microtype}      
\usepackage{lipsum}

\newcommand{\m}{{DCDAN}}

\usepackage{graphics}
\usepackage{amssymb}
\usepackage{subfigure}

\usepackage{algorithm}
\usepackage[noend]{algpseudocode}

\usepackage{amsthm}
\usepackage{amsmath}

\newcommand{\sign}{\operatorname{Sign}}
\newcommand{\EX}{\operatorname{\mathbb{E}}}

\newtheorem{proposition}{Proposition}
\newtheorem{assumption}{Assumption}
\newtheorem{postulate}{Postulate}

\usepackage{tikz}
\usetikzlibrary{bayesnet}
\usetikzlibrary{arrows}
\usepackage{color}
\usepackage{graphicx}
\usepackage{caption}
\usetikzlibrary{backgrounds}

\usepackage{xr}
\usepackage{xcite}
\makeatletter
\newcommand*{\addFileDependency}[1]{
  \typeout{(#1)}
  \@addtofilelist{#1}
  \IfFileExists{#1}{}{\typeout{No file #1.}}
}
\makeatother

\title{Deep Causal Representation Learning for \\Unsupervised Domain Adaptation}

\author{
  Raha Moraffah \\
  Department of Computer Science\\
  Arizona State University\\
  Tempe, AZ 85281 \\
  \texttt{rmoraffa@asu.edu} \\
   \And
   Kai Shu \\
  Department of Computer Science\\
  Arizona State University\\
  Tempe, AZ 85281 \\
  \texttt{kai.shu@asu.edu} \\
   \And
   Adrienne Raglin  \\
  Army Research Lab\\
  2800 Powder Mill Rd\\
  Adelphi, MD 20783 \\
  \texttt{adrienne.raglin2.civ@mail.mil } \\
   \And
 Huan Liu\\
  Department of Computer Science\\
  Arizona State University\\
  Tempe, AZ 85281 \\
  \texttt{hliu@asu.edu} \\
}

\begin{document}

\maketitle

\begin{abstract}
Studies show that the representations learned by deep neural networks can be transferred to similar prediction tasks in other domains for which we do not have enough labeled data. However, as we transition to higher layers in the model, the representations become more task-specific and less generalizable. Recent research on deep domain adaptation proposed to mitigate this problem by forcing the deep model to learn more transferable feature representations across domains. This is achieved by incorporating domain adaptation methods into deep learning pipeline. The majority of existing models learn the transferable feature representations which are highly correlated with the outcome. However, correlations are not always transferable.
 In this paper, we propose a novel deep causal representation learning framework for unsupervised domain adaptation, in which we propose to learn domain-invariant causal representations of the input from the source domain. We simulate a virtual target domain using reweighted samples from the source domain and estimate the causal effect of features on the outcomes. 
 The extensive comparative study demonstrates the strengths of the proposed model for unsupervised domain adaptation via causal representations. 
\end{abstract}
\section{Introduction}
Deep neural networks have had great achievements in different areas such as image classification \cite{Krizhevsky:2012:ICD:2999134.2999257} and object detection \cite{Girshick:2014:RFH:2679600.2679851}. These models usually require huge amounts of training data. However, collecting and annotating datasets are usually labor-intensive. Luckily, there are huge amounts of labeled data available from other domains that can be leveraged. However, distributions of the source (i.e., the learning  model is trained on) and target (i.e., the dataset we wish to apply the trained model on) are usually different, which leads to the failure of those models that transfer the knowledge from source to target domain directly.

Even though deep neural nets can learn transferable representations, studies show that differences between distributions of source and target domains (domain shift) can affect the performance of these models and the representations become less transferable in higher layers of the network \cite{donahue2014decaf}\cite{DBLP:journals/corr/YosinskiCBL14}. Moreover, in many real-world cases where no or few labeled data is available in the target domain, overfitting to the source distribution ensues if the model is trained in a supervised manner on data in both source and target domains \cite{DBLP:journals/corr/abs-1802-03601}. 
Various unsupervised deep domain adaptation methods are proposed to utilize labeled samples from source domain and unlabeled samples from the target domain, and learn a classifier which minimizes the prediction error in the target domain by embedding shallow domain adaptation methods into deep learning pipeline and learning representations that are both predictive and domain invariant \cite{DBLP:journals/corr/Long0J16} \cite{Tzeng:2015:SDT:2919332.2919970} \cite{tzeng2014deep} \cite{long2015learning}. 
%
However, these methods learn feature representations correlated with the outcome.
 \begin{figure}
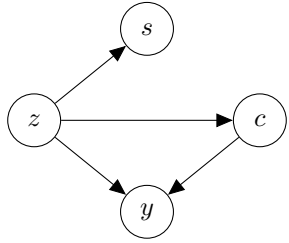

\begin{center}
  \tikz{
 \node[latent] (y) {$y$}; %
 \node[latent,above=of y,xshift=-1.5cm, yshift=-0.5cm] (z) {$z$};
 \node[latent,above=of y,xshift=1.5cm, yshift=-0.5cm] (c) {$c$}; %
 \node[latent,above=of c,xshift=-1.5cm, yshift=-0.5cm] (s) {$s$};
 \edge {c} {y}
 \edge {z} {c}
 \edge {z} {s}
 \edge {z} {y}
}
\caption{A causal diagram: \textbf{$c$} and  \textbf{$s$} are causal and spurious features, \textbf{$z$}  confounders, and \textbf{$y$} prediction outcome.}
\label{example}
\end{center} 
\end{figure}
 Since some correlations can be spurious and therefore not transferable, we propose to learn the \textit{causal} feature representations for unsupervised domain adaptation. 
Causal feature representations of the data are those used to define the structure of the outcome variable rather than the context. 
For instance, consider a picture of a cat. Features such as eyes, ears and shape of the face which are related to the structure of a cat and thus are referred to as causal feature representations, and features such as background of the image are called context feature representations. Figure~\ref{example} illustrates one possible causal graph of the extracted features such as eyes, ears, background and the outcome variable (i.e., an indicator of whether the image is a cat or not). Causal features such as eyes and ears (i.e., $c$) have direct causal effect on the outcome whereas context features (i.e., $s$) are spurious features, i.e. the correlations between them and the outcome variable are due to the existence of $z$, a confounder. The correlations due to confounding variables in the data are misleading and may not be transferable across different target domains. 

The causal mechanism that maps the cause to the effect should not depend on the distribution of the cause \cite{Scholkopf:2012:CAL:3042573.3042635} and causal features are naturally transferable across different domains \cite{Rojas-Carulla:2018:IMC:3291125.3291161}.
For instance, in our example, if the model learns the features related to the structure of cats such as eyes, ears, whiskers and etc., instead of learning the context features, 
these causal features are more transferable across domains. This can be achieved by learning causal relationships between features and the outcome instead of their correlations.




In order to capture the causal relationships of the learned representations on the outcome, 
followng \cite{DBLP:journals/corr/abs-1806-06270}, our framework in Figure~\ref{example} simulates a virtual target domain on top of the representations learned from the source data by re-weighting the samples from the source domain. We show that in this simulated target domain, only representations that are the causes of the outcome variable ($c$) are extracted and all other correlations such as those between $s$ and $y$) are removed. Therefore, the model can learn the representations with highest causal effect on the outcome by measuring the correlations between the outcome and representations of the virtual domain. These representations are then used along with the causal mechanism to perform prediction in a target domain. Learning these weights are embedded into the pipeline of the deep neural net and occurs jointly with parameters of the deep model. This way, the model can learn the representations which are both predictive and invariant across different domains.
Our major contributions are summarized as follows:
\begin{itemize}
    \item We study a novel problem of learning causal representations for unsupervised domain adaptation;
    \item We propose a general framework {\m} to learn causal feature representations and causal mechanisms to make prediction in target domains and show that the learned representations are indeed those with highest causal effects on the target variables; and 
    \item We conduct experiments to demonstrate the effectiveness of the proposed framework for unsupervised domain adaptation by learning causal feature representations.
\end{itemize}
\section{Related Work}


We review research on deep visual domain adaptation and causal inference in domain adaptation and feature learning.


\textbf{Deep visual domain adaptation.}
Despite the achievements of deep neural networks in feature learning, \cite{DBLP:journals/corr/YosinskiCBL14} and \cite{donahue2014decaf} show that the transferability of the features learned decreases by the last layer of the network. Deep domain adaptation addresses the issue by embedding domain adaptation into the pipeline of deep models and learning representations that are both predictive and domain invariant. This is achieved by using several different criteria. For example, \cite{Tzeng:2015:SDT:2919332.2919970}, \cite{Peng2016FinetocoarseKT} and \cite{DBLP:journals/corr/Long0J16} leverage class labels as a guide for transferring knowledge across different domains. \cite{DBLP:journals/corr/Long0J16a}, \cite{DBLP:journals/corr/Long0J16} and \cite{long2015learning} approach the problem by aligning the statistical distribution shifts between source and target domains. To compare the distributions of source and target domains, criteria such as maximum mean discrepancy (MMD) \cite{DBLP:journals/corr/Long0J16a}, \cite{DBLP:journals/corr/Long0J16}, \cite{tzeng2014deep}, \cite{long2015learning}, Kullback-Leibler(KL) divergence \cite{Zhuang:2017:SRL:3154791.3108257}, and correlation alignment (CORAL) \cite{DBLP:journals/corr/SunS16a}, \cite{DBLP:journals/corr/SunFS15} have been used. 
Another line of work focuses on adversarial-based domain adaptation which minimizes the distance between source and target domains through an adversarial objective of the domain discriminator, aiming to encourage domain confusion, including      \cite{Tzeng:2015:SDT:2919332.2919970}, \cite{Ganin:2015:UDA:3045118.3045244}, and \cite{DBLP:journals/corr/0001T16} on visual deep domain adaptation. 



\textbf{Causal inference in domain adaptation.}
Recent work on shallow domain adaptation proposes to learn invariant features for domain adaptation over agnostic target domains. For instance, \cite{Rojas-Carulla:2018:IMC:3291125.3291161} propose a causal framework to identify the invariant features across different datasets and use them for prediction. \cite{PetBuhMei15} propose to find the causal features by exploring the invariance of conditional distribution of the target variable across different domains. 
\cite{DBLP:journals/corr/abs-1806-06270} propose a causal approach to select domain invariant predictors among all predictors and use them to perform domain adaptation across unknown environments. However, they are all designed for shallow domain adaptation and choose useful predictors rather than learning them from the data. 

Causal inference has been also utilized for learning visual features from the data. \cite{Chalupka:2015:VCF:3020847.3020867} proposes a visual causal feature learning framework which constructs causal variables from micro variables in observational data with minimum experimental effort.However, this work requires performing experiments.

\section{Preliminaries of Domain Adaptation}
In this work, given a source domain $D_s = \left\{ (x_i^s, y_i^s) \right\}_{i=1}^{n_s}$ with ${n_s}$ labeled samples, we predict the labels for an unlabeled target domain $D_t = \left\{x_i^t \right\}_{i=1}^{n_t}$ by leveraging samples from the source domain to minimize prediction errors. Let \textit{P} and \textit{Q} denote the distributions of the source and target domains, respectively.  \textit{P(X, Y)} and \textit{Q(X, Y)} are the joint distributions of the inputs and outcomes for source and target domains. In general, these two joint distributions are different, i.e., $\textit{P(X, Y)} \neq \textit{Q(X, Y)}$. Following the traditional setting~\cite{DBLP:journals/corr/abs-1802-03601}, we have two basic assumptions: (i) $P(Y_S|X_S) = P(Y_T|X_T)$, which means the conditional distribution of the outcome given the data remains the same across different environments; and (ii) $\textit{P(X)} \neq \textit{Q(X)}$, indicating that the difference between the joint distribution of the inputs and outcomes originates from the difference between the marginal distributions of the inputs. Moreover, we assume that the difference between marginal distributions of the inputs comes from a drift in the feature space of the problem (or covariate shift).
Specifically, we study the problem of
%
constructing a deep neural network that learns transferable representations, Z, for which the conditional probability of the outcomes 
remains the same across different domains and learns a classifier ($\theta$ (.)) such that $\theta$ maps the learned representations to the outcome and the target loss $\epsilon (\theta) = \Pr_{(x, y)\sim Q} [\theta(x) \neq y]$ is minimized. 

\section{Proposed Framework - {\m}}
\begin{figure}
\centering
\includegraphics[width=100mm,scale=1.4]{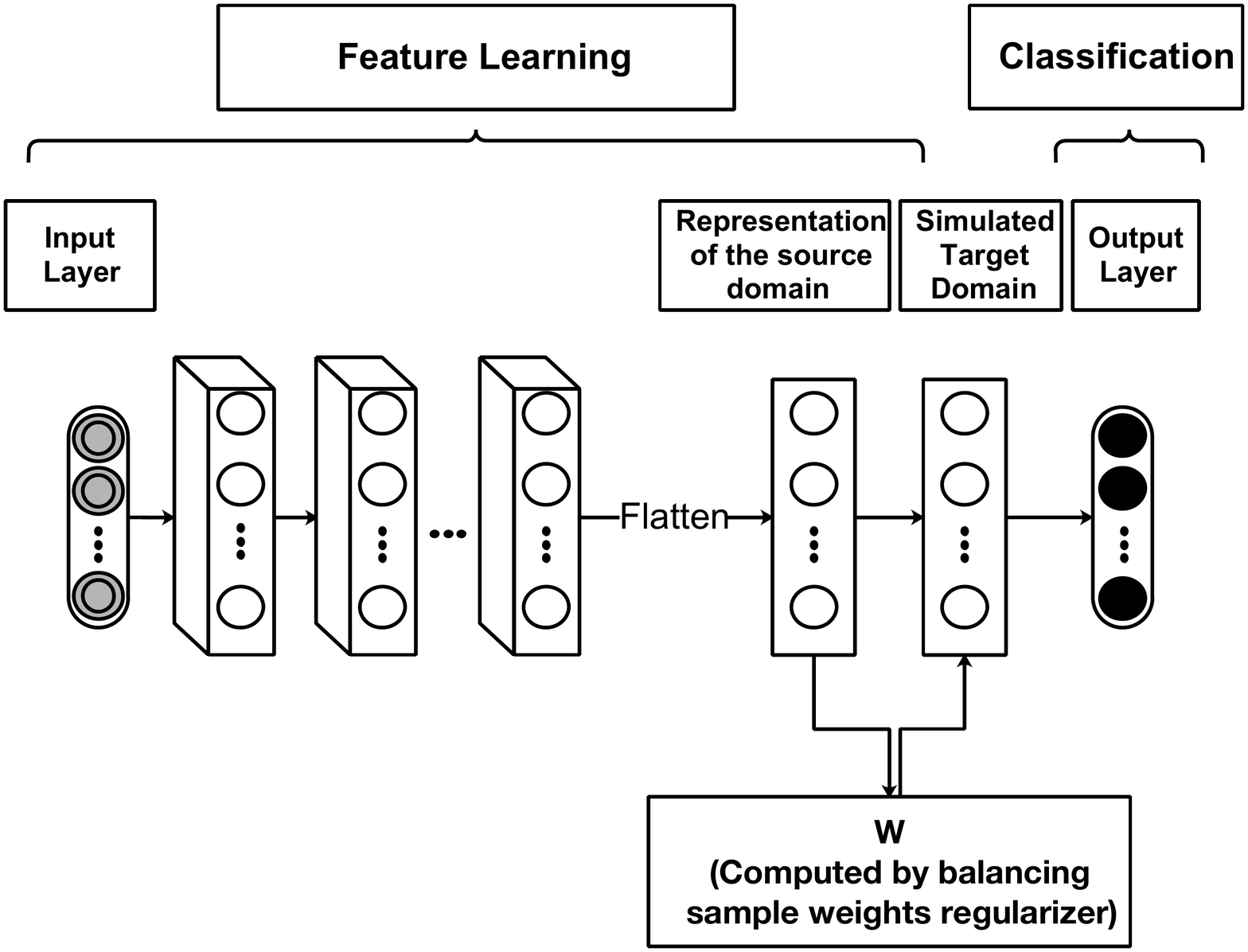}
\caption{An overview of {\m} that learns causal representation of the data for prediction.} 
\label{fig:framework}
\end{figure}

We propose a Deep Causal Representation learning framework for unsupervised Domain Adaptation {(\m)} to learn transferable feature representations for target domain prediction. A deep neural network combines feature extraction and a classifier which learns the highly correlated feature representations with the outcome. However, some of these correlations could be due to biases in the data, e.g., confounding bias. We aim to remove spurious correlations learned by the model by measuring the causal effects of the representations on the outcome.
{\m} consists of a regularization term which learns the balancing weights for the source data by balancing the distribution with respect to feature representations learned from the data. These weights are designed in a way to help the model capture the causal effects of the features on the target variable instead of their correlations.  Moreover, our model includes a weighted loss function of deep neural net, where the weight of each sample comes from the regularization term and the loss function is in charge of learning predictive domain invariant features as well as a classifier which maps the learned representations to the output, or a causal mechanism. 
By embedding the sample weights of the learning component into the pipeline of the model and jointly learning these weights with the representations, we can benefit from the deep model to learn causal features that are both transferable and good predictors for the target. The framework is shown in Figure ~\ref{fig:framework}.

\subsection{Balancing Sample Weights Regularizer}
As discussed in the previous section, in order to learn the representations with causal effect on the outcome (i.e. causal feature representations) and reduce the confounding bias, we need to force the deep neural net to learn the causal relationships instead of the correlations. 
To do so, following \cite{DBLP:journals/corr/abs-1806-06270}, we reweight samples in the source domain with sample weights which enable us to capture the causal effect of each learned feature on the outcome variable by controlling the effect of all other features in the learned representation. Variable balancing techniques are often used in causal effect estimation from observational data where the distributions of the covariates are different between treatment and control groups due to the non-random assignment of the treatments to the units. In order to get an unbiased estimation of the causal effect of the treatment variables, one approach is to balance distributions of the treatment and control groups by applying balancing weights on the samples. One way to learn these weights is to characterize the distributions of the treatment and control groups by their moments and learn the weights W as following: 

\begin{equation*}
\begin{split}
& W = \underset{W}{\text{argmin}}
\Vert \frac{\sum_{i:T_i = 1}w_i\cdot x_i}{\sum_{i:T_i = 1} w_i} - \frac{\sum_{i:T_i = 0}w_i\cdot x_i}{\sum_{i:T_i = 0} w_i}\Vert_{2}^{2} \\
\end{split}
\end{equation*}
where T is the treatment variable and T = 1 and T = 0 represent the treatment and control groups respectively.

It is shown that by considering each feature learned in the set of feature representations as a treatment variable and learning weights to balance the distribution of the source data with respect to every learned feature representation, we can learn a new domain, in which only causal feature representations are correlated with the outcome. These new representations then can be used to estimate the causal contributions of source representations on the outcome.
The sample weights can be learned by minimizing the below function: 


\begin{equation*}
\begin{split}
\small
G=\sum_{i=1}^{\vert Z \vert} \Vert \frac{Z^T_{.,-i} \cdot (W \circ I(Z_{.,i}))}{W^T \cdot I(Z_{.,i})} 
- \frac{Z^T_{.,-i} \cdot (W \circ (1 - (I(Z_{.,i}))))}{W^T \cdot (1 - I(Z_{.,i}))}\Vert_{2}^{2}
\normalsize
\end{split}
\end{equation*}
where $W \in \mathbb{R}^{n_s \times 1}$ denotes a vector of sample weights, Z = h(X) represents the feature representations extracted from the deep model, $Z_{.,i}$ is a vector of i-th feature representation of all samples, $Z_{.,-i}$ is the set of all features representations except the i-th ones, $\circ$ refers to the Hadamard product and I is and indicator matrix that indicates whether the feature exists in data samples (i.e. the entry is equal to one, which means the samples belong to the treatment group) or does not exist (the entry is equal to zero and, which means data samples belong to the control group). In order to create the Indicator matrix, we binarize the representations by leveraging the methods proposed in \cite{DBLP:journals/corr/CourbariauxBD15} for binarizing neural networks. In order to binarize the representations, two different possible approaches can be used. The first function is deterministic: 

\[ 
x^b = \sign(x)= \left\{
\begin{array}{ll}
      +1 & x \geq 0, \\
      0 & \text{otherwise,}\\
\end{array} 
\right. 
\]
where $x^b$ is the binarizaed version of the real-valued variable x. This function is easy to implement and is shown to work well in practice. The second function is stochastic:
\[ 
x^b = \sign(x)= \left\{
\begin{array}{ll}
      +1 & \text{with probability $p = \sigma(x) $,}\\
      0 & \text{with probability $1-p $,}\\
\end{array} 
\right. 
\]
where $\sigma$ is the \textit{"hard-sigmoid"} function defined as:

\begin{equation*}
\begin{aligned}
\sigma(x) = \operatorname{clip}(\frac{x+1}{2}, 0 , 1) = \max(0, \min(1, \frac{x+1}{2}))
\end{aligned}
\end{equation*}

This stochastic function 
makes the overlap assumption (\ref{overlap}) more plausible than a deterministic function. 
Since the derivative of both binarization functions are almost zero everywhere during the back-propagation, following Hinton 2012's lectures and \cite{DBLP:journals/corr/CourbariauxBD15}, we use ``straight-through estimator" for back propagation. 

\par

\subsection {Deep Causal Domain Adaptation}

We explore the idea of using a  sample reweighting regularizer for learning transferable features in deep networks. 
Convolutional Neural Networks (CNN) are  widely used deep frameworks in computer vision \cite{Krizhevsky:2012:ICD:2999134.2999257}  and achieved great performance in a variety of tasks. However,  
as discussed earlier, transferring these models to new domains, layers of the CNN model become less transferable at more task-specific layers. Therefore, it requires vast amounts of labeled data in the target domain to fine-tune the model and avoid over-fitting. Unsupervised domain adaptation frameworks such as \cite{long2015learning} and \cite{tzeng2014deep} leverage unlabeled data in the target domain and learn the more transferable representations in the source domain to the target domain. They learn transferable feature representations highly correlated with the outcome. Relying on correlations limits their performance since correlations do not necessarily exist in other domains, thus, may not be transferable. If we find only transferable feature representations, the performance on the target domain can be further improved.
We propose to learn causal representations of the input for which conditional distributions of the outcomes (a.k.a. causal mechanisms) remain the same across different domains even if the distributions of the inputs change \cite{Scholkopf:2012:CAL:3042573.3042635}. this can be achieved by reweighting the learned representations with sample weights learned by a balancing sample regularizer. Reweighted samples play the role of a virtual target domain in which only representations with causal contributions on the outcome are correlated with the target and spurious correlations between two variables that do not truly exist and are due to the confounders are removed. By doing so, we force the model to learn features with highest causal contributions on the outcome. 

We implement {\m} on the Resnet-50 architecture~\cite{he2016deep} which can be easily replaced with any other convolutional neural network framework. Resnet-50 is used as a backbone for a variety of tasks in computer vision such as deep transfer learning. It consists of 5 stages with convolutions and identity blocks. Each convolution and identity block consist of 3 convolution layers. Resnet-50 leverages the concept of skipping connections, which proposes to add the original input to the output of the convolution block while stacking the convolution layers together, to reduce the risk of vanishing the gradients during back-propagation.
The empirical risk of CNN can be written as:

\begin{equation}\label{eq:2.1}
\min_{\Theta} \frac{1}{n_s} \sum_{i = 1}^{n_s} J(\Theta(x_i^s), y_i^s) 
\end{equation}
where \textit{J} is the cross-entropy loss function, $\Theta$ parameters of CNN, and $\Theta(x_i^s)$ the conditional probability that CNN assigns $x_i^s$ to label $y_i^s$.

To learn invariant causal feature representations among different domains 
by using a balancing sample reweighting regularizer, we propose to jointly learn the sample weights from the data, reweight the representations by these these weights and minimize the loss of prediction for the the reweighted samples. This can be done by reweighting the loss of each sample with its corresponding weight and minimizing the weighted empirical loss of CNN and the balancing regularizer G simultaneously. This way, we can reduce the bias in the correlations learned by the model and learn the features with highest causal effect on the output that are also informative for prediction.
Therefore, we embed the balancing sample reweighting regularizer into the CNN framework as: 

\begin{equation}
\begin{aligned}
\label{EQ:main1}
& \underset{\Theta , W}{\text{min}}
\frac{1}{n_s} \sum_{i = 1}^{n_s} w_i J(\Theta(x_i^s), y_i^s) +  
\lambda_1 G \\
& \text{subject to}~ W \geq 0, \; \Vert W\Vert_2^2 \leq \lambda_2 ,   \sum_j^{\vert h(X) \vert}w_j-1)^2 \leq \lambda_3
\end{aligned}
\end{equation}
where $\sum_{i = 1}^{n_s} w_i J(\Theta(x_i^s), y_i^s)$ is the weighted cross-entropy loss function, representing the weighted empirical loss of CNN model, G the balancing sample reweighting regularizer, $W \in \mathbb{R}^{n_s \times 1}$ a vector of sample weights, $W \geq 0$ ensures all the weights are non-negative, 
$\Vert W\Vert_2^2 \leq \lambda_2$ tries to reduce the sample variance and $(\sum w_j-1)^2 \leq \lambda_3$ avoids all sample weights from being zero. 

To optimize {\m}, we update $\theta$ and W iteratively using mini-batch SGD. 
A detailed algorithm of the optimization framework can be found in the supplementary material.

\subsection{Theoretical analysis}
In this section, we explain the key assumption of our proposed method and provide some analysis on the reasons why the method learns the causal features (i.e. the features with highest causal contributions on the target variable). In order for our method to work, we need to make the overlap assumption, which is a common assumption in causal inference literature \cite{DBLP:journals/corr/abs-1806-06270}. Overlap assumption ensures that for each data instance in the treatment group, a counterpart from control group can be found.

\begin{assumption}\label{overlap}
[Overlap] For any variable $T_{i}$, where $T_{i}$ is the treatment variable for i-th sample in the data, $0<\Pr(T_i = 1|X = x_i)<1$ for any $x_i$ in the dataset.
\end{assumption}

\begin{proposition} 
\label{prop1}
Feature representations learned by {\m} 
have highest causal contributions on the outcome.
\end{proposition}
Proof of this proposition is in the supplementary material.

\begin{postulate} [Independence of mechanism and input]\label{pos1}

Following \cite{Scholkopf:2012:CAL:3042573.3042635}, we assume that the causal mechanism is "independent" of the distribution of the cause. In other words, $P(E|C)$ contains no information about $P(C)$ where E is the effect (i.e. outcome) and C is the cause. This indicates that changes in $P(C)$ has no influence on the mechanism after it is learned.

\end{postulate}
Postulate \ref{pos1} implies that once the causal mechanism (i.e. $P(E|C)$) is learned, we can assume that it remains the same even when the distribution of the input (i.e. $P(x)$) and therefore distribution of causes (i.e. $P(C)$) changes. Therefore, we can address the covariate shift problem \cite{Sugiyama:2012:MLN:2209761}, where $P(Y|X)$ remains the same across different environment while $P(X)$ changes, by learning the causal features from the input and a functions that maps those feature to the outcome.
\section{Experiments}



\begin{figure*}[t!]
\centering
\subfigure[]{\includegraphics[width=10pc, height = 10pc]{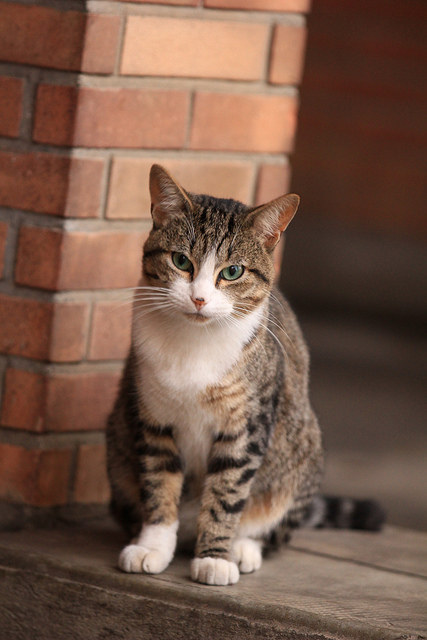}\label{pic1}}\quad%
\subfigure[]{\includegraphics[width=10pc, height = 10pc]{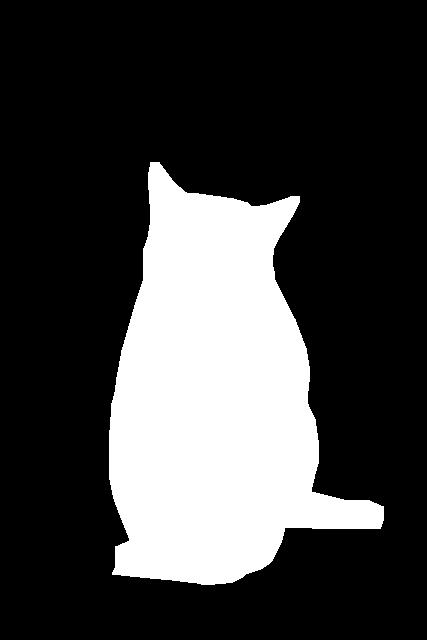}\label{pic2}}\quad
\subfigure[]{\includegraphics[width=10pc, height = 10pc]{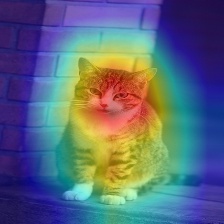}\label{fig:images3.3}}\quad\\
\caption{An example of samples in dataset constructed to perform \textbf{(EQ2)} and heat-map generated by {\m}. Figure \ref{pic1} shows a sample image from the data, Figure \ref{pic2} shows the ground truth for causal features of figure \ref{pic1} extracted from VQA-X dataset and figure \ref{fig:images3.3} shows the heat-map generated by {\m} for the causal feature representations}
\label{fig:images3}
\end{figure*}


Our experiments are designed to evaluate the effectiveness of the proposed {\m} with the following questions:
\begin{itemize}
    \item \textbf{EQ1}: How is {\m} compared to existing unsupervised deep domain adaptation frameworks?
    \item \textbf{EQ2}: Are the feature representations learned by {\m}, causal features for predicting outcomes?
    \item \textbf{EQ3}: How does varying the causal regularizer and other hyperparameters affect the classification performance of {\m}?
\end{itemize}
We introduce the datasets and representative state-of-the-art deep domain adaptation frameworks, then compare the performance of {\m} for an object recognition task to answer \textbf{EQ1}. For \textbf{EQ2},  we investigate the ability of {\m} to learn and extract causal features from the data. For \textbf{EQ3}, we perform experiments by varying all hyperparametes of the model and report their performance.
\subsection{Datasets}\label{dataset}

To answer \textbf{EQ1}, following the convention for domain adaptation study, we use \textit{Office-31} and \textit{Office-10}+\textit{Caltech-10}, two of the widely-adopted, publicly available benchmark datasets.  Office-31 consists of 4,652 images within 31 categories collected from three distinct domains: Amazon ($A$), which contains images downloaded from amazon.com, Webcam ($W$) and DSLR ($D$), which are images taken by web camera and digital SLR camera in
an office with different environment variation. For a comprehensive comparison, following~\cite{long2015learning}, we evaluate the performance of all models by considering all possible pairs among the three domains: $A\rightarrow W$, $D\rightarrow W$, $W\rightarrow D$,  $A\rightarrow D$, $D\rightarrow A$, and $W\rightarrow A$. Office-10 + Caltech-10 dataset consists of 10 common categories shared by the Office31 and Caltech-256 ($C$). Thus, for all 4 domains ($A$, $W$, $D$, and $C$), we conduct evaluations on all remaining possible pairs involving $C$: $A\rightarrow C$, $W\rightarrow C$, $D\rightarrow C$,  $C\rightarrow A$, $C\rightarrow W$, and $C\rightarrow D$.
\begin{table*} [htbp!]
\small
\vspace{-0.2cm}
\centering \caption{The Prediction Performance on Domain Adaptation on \texttt{Office-31}}
\begin{tabular}{|l|c|c|c|c|c|c|c|}
\hline
Method & $A\rightarrow W$ & $D\rightarrow W$ & $W\rightarrow D$ & $A\rightarrow D$ & $D\rightarrow A$ & $W\rightarrow A$ &Average   \\
\hline
\hline
ResNet-50 & 69.685&97.610 & 99.405&71.485 & 63.080 & 61.448& 77.118\\
\hline
DDC & 77.987& 96.981& 100.000& 81.526&65.246 & 64.004& 80.957\\
\hline
DAN &82.000 & 97.000&100.000 &83.000 & 66.000& 65.000&82.166 \\

\hline
DeepCoral &77.800 & 97.700& 99.700& 81.500& 64.600& 64.000& 80.883\\
\hline
DANN & 82.000& 96.900& 99.100& 79.700& 68.200& 67.400& 82.216\\
\hline
HAFN & \textbf{82.900}& 98.100& 99.600& 83.700& \textbf{69.700}& 68.100& 83.683\\
\hline
{\m} & 81.000 & \textbf{99.000}& \textbf{100.000}& \textbf{86.000}& 69.000& \textbf{70.000}& \textbf{84.166}\\
\hline
\end{tabular} \label{tab:performance}
\end{table*}

\begin{table*} [htbp!]
\small
\vspace{-0.2cm}
\centering \caption{The Prediction Performance on Domain Adaptation on \texttt{Office-10+Caltech-10}}
\begin{tabular}{|l|c|c|c|c|c|c|c|}
\hline
Method &$A\rightarrow C$ & $W\rightarrow C$ & $D\rightarrow C$ & $C\rightarrow A$ & $C\rightarrow W $& $C\rightarrow D$ &Average   \\
\hline
\hline
ResNet-50 & 86.375 &89.671 &87.978 &93.423 &93.559 & 93.631&90.772\\
\hline
DDC &91.184 &89.670 &  89.586&94.989 & 95.932&\textbf{96.815} &93.029 \\
\hline
DAN &91.000 &89.000 &87.000 & 95.000& 96.000& 96.000&92.333 \\
\hline
DeepCoral &90.293 & 88.691& 87.529 &\textbf{95.198} & 95.593& 96.178& 92.247\\
\hline
DANN & 91.451 &87.000 & 84.862&  94.000& 94.000& 92.000&90.552 \\
\hline
HAFN & 90.115 & \textbf{91.629} &\textbf{95.302} &91.718 & 95.254& 92.356 & 92.729\\
\hline
{\m} &\textbf{92.000} & 90.000 &91.000 &94.000 & \textbf{96.000}& 96.000& \textbf{93.166}\\
\hline
\end{tabular} \label{tab:performance1}
\end{table*}

To answer \textbf{EQ2}, a dataset with causal features ground truths is needed. However, obtaining ground truths for causal features of objects is a difficult task since existing datasets for object recognition do not include the causal features of the targets and only contain the ground truths for the labels of the images in the datasets. Therefore, to answer \textbf{EQ2}, we construct a dataset with reliable ground truth for causal features corresponding to target variables by utilizing a subset of Visual Question Answering Explanation (VQA-X) dataset proposed in~\cite{huk2018multimodal}, where a set of images extracted from MSCOCO dataset\footnote{http://cocodataset.org}
along with a set of questions, answers, visual and textual explanations for questions are provided by human annotators. To construct the dataset for our study, we extract a subset of VQA-X dataset that contains only single objects with their corresponding labels from MSCOCO dataset. Our dataset consists of actual images, their labels and the visual explanations of the target variable which represents the causal feature representations of the target. These visual explanations are given by heatmaps provided by human experts.
Figures~\ref{pic1} and \ref{pic2} show one example of the data. 

\subsection{Compared Baseline Methods}\label{sec:baseline}
    In this section, we briefly introduce the representative state-of-the-art baseline methods for deep domain adaptation:
\begin{itemize}
    \item ResNet-50~\cite{he2016deep}: It is a state-of-the-art convolution neural network model for image classification. It utilizes a deep residual neural network structure which introduces the identity shortcut connect to skip layers automatically to avoid overfitting for deep neural networks. 
    \item DDC~\cite{tzeng2014deep}: DDC is a deep domain confusion model that aims to maximize the domain invariance by adding an adaptation layer in convolution neural networks with a single-kernel maximum mean discrepancies (MMD) regularization proposed in \cite{Scholkopf2012}.  
    \item DAN~\cite{long2015learning}: DAN is a deep adaptive neural network model for unsupervised domain adaptation. It reduces the domain discrepancy via optimal multi-kernel selection for mean embedding matching.
    \item Deep CORAL~\cite{DBLP:journals/corr/SunS16a}: Deep CORAL is an unsupervised deep domain adaptation framework, which learns a non-linear transformation that aligns the second-order statistics between the source and target feature activations in deep neural networks.
    \item DANN~\cite{Ganin:2015:UDA:3045118.3045244}: DANN is an adversarial representation learning framework in which one classifier aims to distinguish the learn source/target features while the another feature extractor tries to confuse the domain classifier. The minimax optimization is then solved through a gradient reversal layer.
    \item HAFN ~\cite{DBLP:journals/corr/abs-1811-07456}: Hard Adaptive Feature Norm is a variantl of 
    AFN, a non-parametric Adaptive Feature Norm framework for unsupervised domain adaptation, based on adapting feature norms of source and target domains to achieve symmetry over a large number of values.
\end{itemize}

\subsection{Classification Performance Comparison}
To answer \textbf{EQ1}, we compare {\m} with aforementioned representative methods. Following existing works on deep domain adaptation, we utilize accuracy as the evaluation metric. 
For baseline methods, we follow standard evaluation mechanism for unsupervised domain adaptation and use all source instances with labels and all target instances without labels~\cite{long2015learning}. We implement all the baselines using PyTorch~\footnote{https://pytorch.org/}. In addition, we evaluate all compared approaches through grid search on the hyperparameter space, and report the best results. For MMD-based methods (i.e. DDC and DAN), we use Gaussian kernels. The experimental results are shown in Table~\ref{tab:performance} and Table~\ref{tab:performance1}. From the tables, we make the following observations:
\begin{figure}[hb!]
\centering
\subfigure[{\m}]{\includegraphics[width=9pc, height = 10pc]{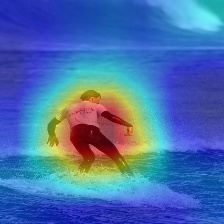}}\quad
\subfigure[Resnet-50]{\includegraphics[width=9pc, height = 10pc]{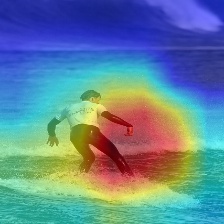}}\quad\\
\caption{Heatmaps generated by {\m} and Resnet-50 on a subset of VQA-X data.}
\label{fig:images}
\end{figure}
\begin{itemize}
    \item Without access to data in the target domain, {\m} still outperform the baselines in many cases for both datasets, which validates that causal feature representations are helpful for learning transferable features across domains.
    \item {\m} significantly outperforms Resnet-50, the only baseline that 
    does not use any information from the target domain, which suggests that {\m} can perform unsupervised domain adaptation. 

\end{itemize}
\begin{figure*}[ht!] 
\centering
\subfigure[Accuracy vs. $\lambda_1$]{\includegraphics[width=12pc, height = 12pc]{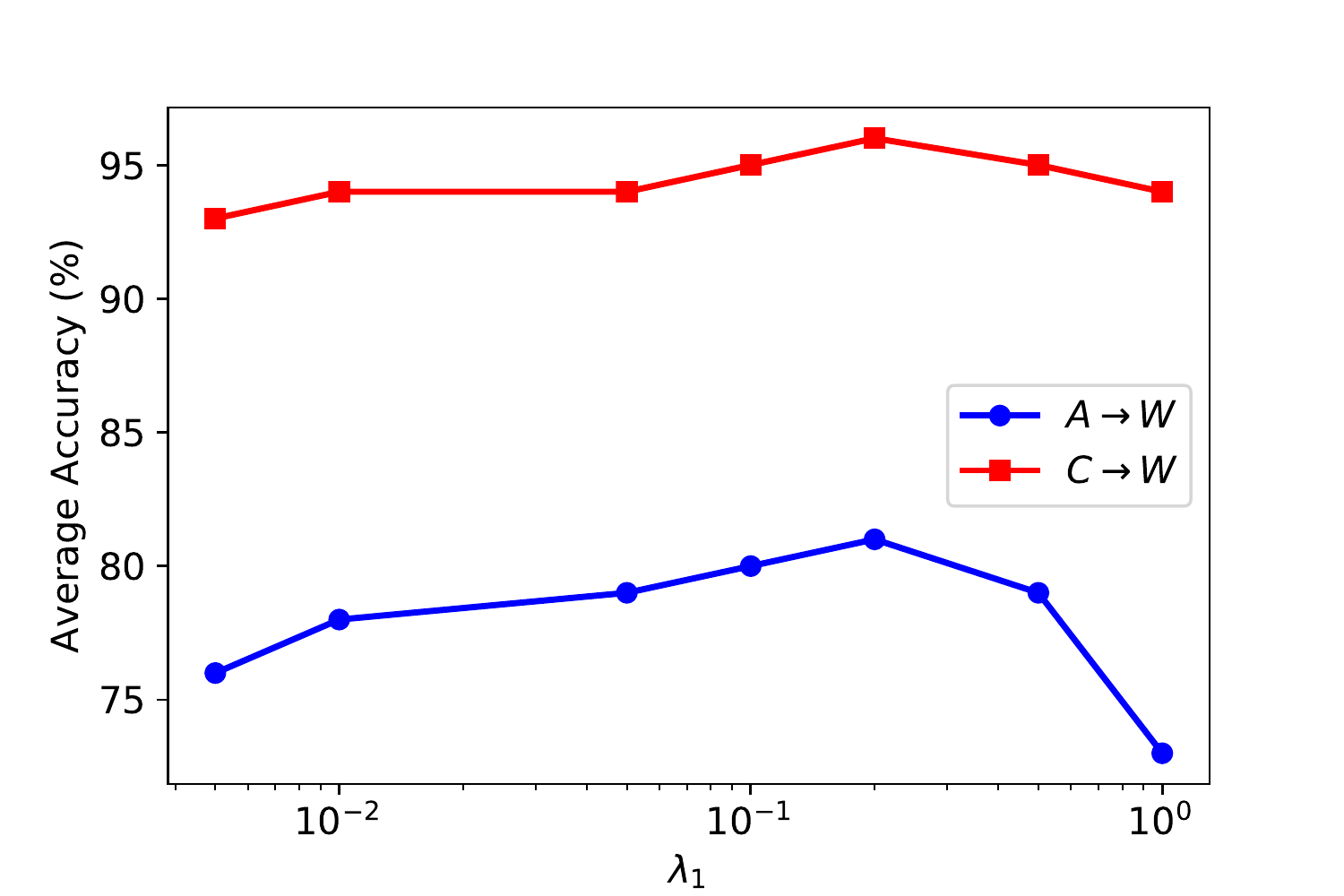}\label{fig2}}\quad%
\subfigure[Accuracy vs. $\lambda_2$]{\includegraphics[width=12pc, height = 12pc]{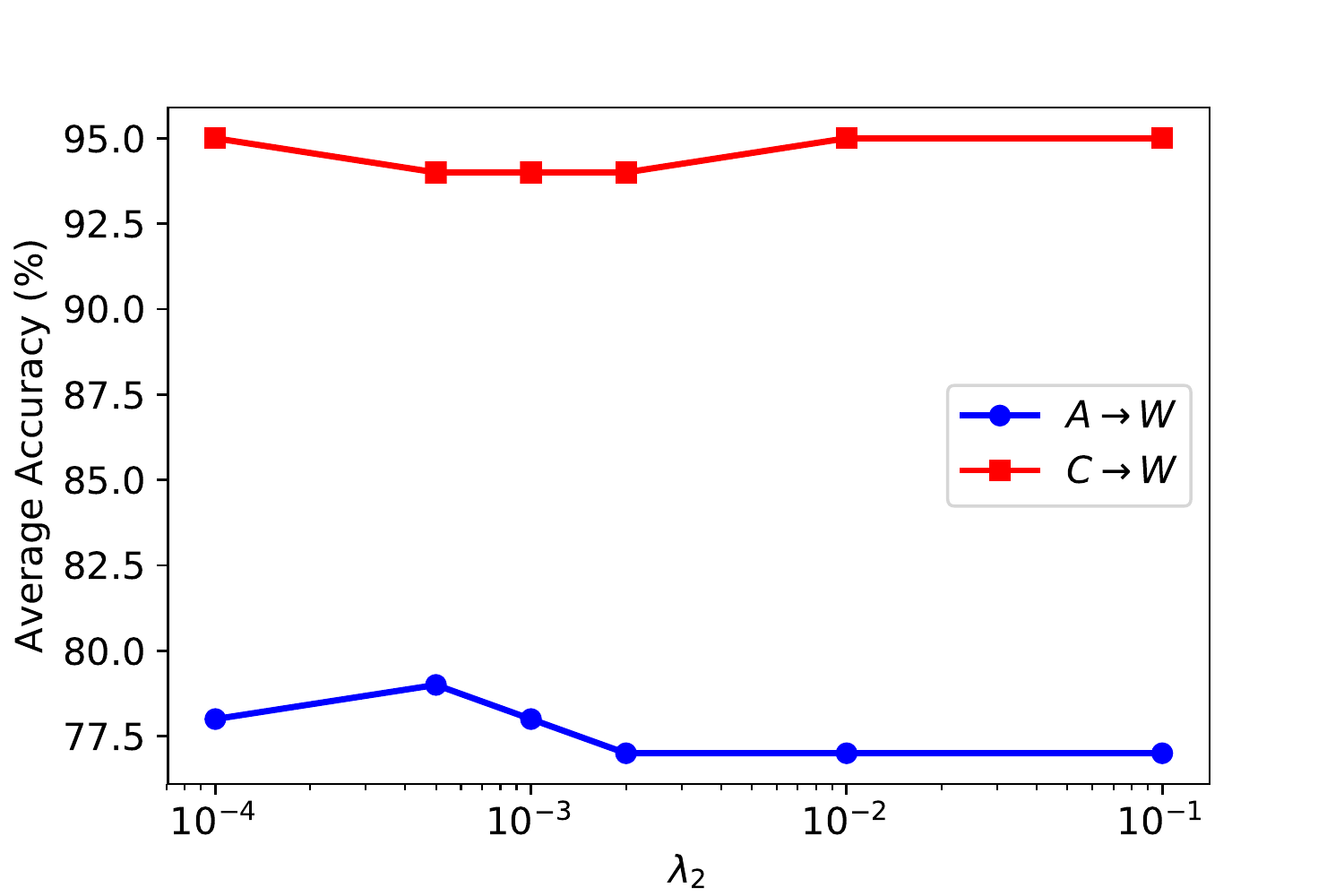}\label{fig3}}\quad
\subfigure[Accuracy vs. $\lambda_3$]{\includegraphics[width=12pc, height = 12pc]{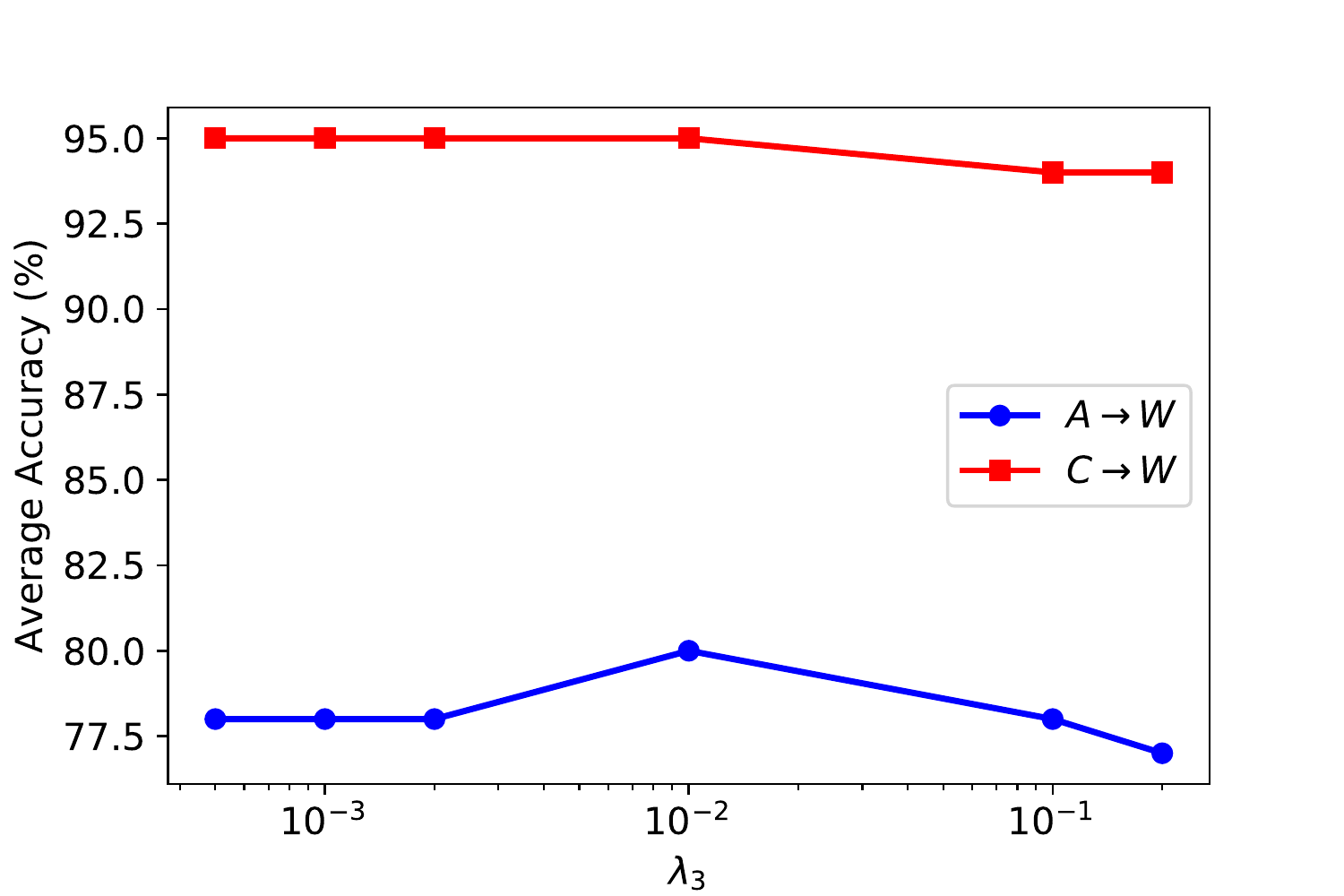}\label{fig4}}\quad\\
\caption{Accuracy of {\m} with varying hyperparameters $\lambda_1$, $\lambda_2$ and $\lambda_3$ on tasks \textbf{$A \rightarrow W$ } and 
    \textbf{$C \rightarrow W$ }}
\label{fig:images2}
\end{figure*}

\subsection{Causal Feature Evaluation}

In this subsection, to answer \textbf{EQ2}, we evaluate the performance of {\m} for discovering causal features automatically from the data. It is worth mentioning that all of the baselines used in our first experiment are  designed for improving the performance of image classification in the target domain and none of them are initially proposed to discover interpretable causal features. However, due to the nature of our approach, it is expected to be able to learn more interpretable features from the data by looking for the features that belong to the structure of the object rather than the context.
In order to show the effectiveness of our model in learning causal features, we propose to run {\m} and a pretrained Resnet-50 on a subset of VQA-X dataset \cite{huk2018multimodal} as described in "Datasets" section, extract their feature representations heatmaps and compare them to the visual ground truths provided in VQA-X.

In order to make a fair comparison, we fine-tune our model on a small subset of single-object images in Imagenet dataset \cite{DBLP:journals/corr/RussakovskyDSKSMHKKBBF14} and extract the feature representation heatmaps from the fine-tuned {\m} and pre-trained Resnet-50 using the method proposed in \cite{selvaraju2017grad}. Figure \ref{fig:images3.3} shows an example generated heatmap by {\m}. To compare the generated heatmaps, Following traditional settings, we use Rank correlarion as our evaluation metric. 
Following \cite{das2017human}, to calculate the Rank Correlation, we first scale both the ground truths and heatmaps generated by the models to 14x14, rank the pixels according to their spatial attention and compute the correlation between two ranked lists.
Our experiment shows that the rank correlation for {\m} is \textbf{0.4501} whereas the rank correlation for the pre-trained Resnet-50 is \textbf{0.4077}, which demonstrates the effectiveness of {\m} on learning causal representations. Figure \ref{fig:images} shows an example of the generated heatmaps by {\m} and pre-trained Resnet-50. As seen in Figure \ref{pic2}, the feature representation heatmaps generated by {\m} are more focused on the causal features of the object, whereas the features learned by Resnet-50 can contain both causal and context features.

\subsection{Parameter Analysis}
To answer \textbf{(EQ3)}, we investigate the effect of hyperparameters of the model on the performance (i.e. accuracy of the model) and report the results in Figure~\ref{fig:images2}. The performance is reported on tasks \textbf{$A \rightarrow W$ } and \textbf{$C \rightarrow W$}. To be more specific, Figure~\ref{fig2} illustrates the performance of the balancing sample weights regularizer ($\lambda_1$) on classification performance of {\m} as it varies in the $\lambda_1 \in \{ 0.005, 0.01, 0.05, 0.1, 0.2 , 0.5, 1\}$. We observe that the accuracy of {\m} first increases and then decreases as $\lambda_1$ varies in the mentioned range. This further confirms that a good trade-off between causal loss and classification loss can result in learning more transferable features.
Figure~\ref{fig3} shows the effect of $\lambda_2 \in \{ 0.0001, 0.0005,0.001,0.002, 0.01, 0.1\}$ the hyperparameter which controls the sample variance, on the accuracy of {\m}.
We also report the effect of $\lambda_3 \in \{ 0.0005,0.001, 0.002, 0.01,0.1,0.2\}$ which is designed to prevent all sample weights from being zero in Figure~\ref{fig4}. To measure the effect of $\lambda_3$ on the performance.

\section{Conclusion}
In this paper, we propose {\m}, a novel Deep Causal Representation learning framework for unsupervised domain adaptation to generate more transferable feature representations by extracting causal feature representations instead of only considering the correlations. In order to learn the causal representations a virtual target domain consists of reweighted samples is generated. These sample weights are learned concurrently with the feature representations and in the pipeline of the deep model. Experiments demonstrate the effectiveness of our model for both classification performance and learning causal feature representations.



\clearpage
\section{Proof of Proposition 1}

\textit{Feature representations learned by {\m} are the features with highest causal contributions on the outcome.}

\begin{proof}

Following \cite{DBLP:journals/corr/abs-1806-06270}, it can be proved that 1) reweighted feature representations in the simulated target domain are independent of each other and 2) In the new target domain, only causal feature representations are correlated with the outcome and the possible spurious correlations between the context features and outcome vanish. To be more specific, we have:
\begin{equation*}
\begin{split}
 &\Pr(Y_T = y | S_T = s)  \\
= &\EX_{C_T}(\Pr(Y_T = y | C_t, S_T = s) | S_T = s)\\
= &\EX_{C_T}(\Pr(Y_T = y | C_t) | S_T = s) \\
= &\Pr(Y_T = y)
\end{split}
\end{equation*}
where T denotes the target domain, C represents all causal features, S denotes all spurious variables and Y is the outcome variable.

This shows that by simulating a target domain using the balancing sample weights, correlation between the simulated representations for virtual target domain and the outcome variable can be measured to estimate the unbiased causal contribution of representations on the target and using these virtual representations in the pipeline of the deep model, we can learn the feature representations with highest causal contributions on the data instead of the using correaltions. In other words, features with high spurious correlations are not learned.
\end{proof}
\section{Optimization of {\m}}
Algorithm \ref{algo1} explains the optimization procedure of the proposed framework. {\m} utilizes an alternating optimization approach to solve the optimization problem defined in Eq (2). To be more specific, {\m} updates $\theta$ and W iteratively using mini-batch stochastic gradient descent. 

\begin{figure*}
\centering
\subfigure[Heatmap generated for an instance of pizza class by {\m}]{\includegraphics[width=10pc, height = 10pc]{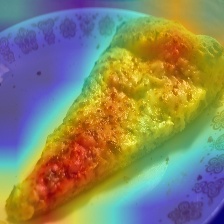}}\quad%
\subfigure[Heatmap generated for an instance of bird class by {\m}]{\includegraphics[width=10pc, height = 10pc]{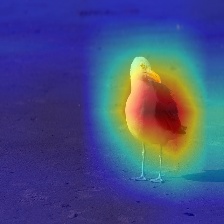}}\quad
\subfigure[Heatmap generated for an instance of dog class by {\m}]{\includegraphics[width=10pc, height = 10pc]{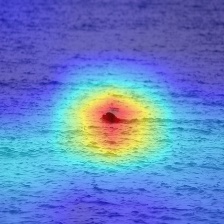}}\quad\\
\subfigure[Heatmap generated for an instance of pizza class by Resnet-50]{\includegraphics[width=10pc, height = 10pc]{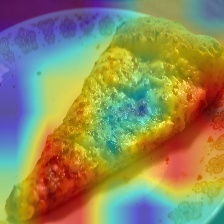}}\quad
\subfigure[Heatmap generated for an instance of bird class by Resnet-50]{\includegraphics[width=10pc, height = 10pc]{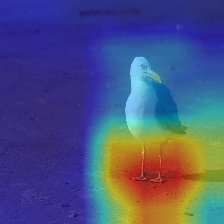}}\quad
\subfigure[Heatmap generated for an instance of dog class by Resnet-50]{\includegraphics[width=10pc, height = 10pc]{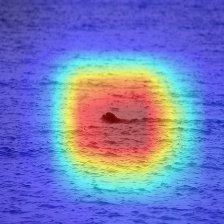}}\quad
\caption{Heatmaps generated by {\m} and Resnet-50 on a subset of VQA-X data. First row shows the feature representation heatmaps generated by our propose {\m} and sencond row shows the heatmaps generated by pre-trained Resnet-50}
\label{fig:images}
\vspace{10cm}
\end{figure*}
\begin{algorithm}
\caption{Deep Causal Domain Adaptation Network algorithm}\label{algo1}
\begin{algorithmic}[h!]
\Require Matrix of input images X and ground truth labels Y  
\Ensure Updated parameters of the model W and ~$\theta$
\State Initialize the iteration variable $t \leftarrow 0$
\State Initialize parameters $W^{(0)}$ and $~\theta^{(0)}$
\State Calculate the loss function with ($W^{(0)}$, $~\theta^{(0)}$) as stated in Eq (2) in the paper .
\Repeat
\State $t \leftarrow t + 1$
\State $\theta^t \leftarrow$ update $\theta$ using Stochastic Gradient Descent while W is fixed
\State $W^t \leftarrow$ update W using Stochastic Gradient Descent while $\theta$ is fixed
\State Calculate loss function with $(W^t, \theta^t)$

\Until {Loss function converges or maximum iteration is reached}
\State {}
\Return W, $\theta$

\end{algorithmic}
\end{algorithm}

\section{Additional Case Studies for Learnig Causal Feature Representations}
In this section, we provide more case studies for "Causal Feature Evaluation" section of the paper. Both {\m} and Resnet-50 are trained according to the settings explained in "Causal Feature Evaluation" section on a subset of VQA-X dataset \cite{huk2018multimodal} as described in "Datasets" section. Figure \ref{fig:images} shows some examples of the heatmaps generated by both our proposed framework and Resnet-50. Heatmaps generated by {\m} are more focused on the regions which blong to the structure of the outcome (i.e. causal features), whereas the features learned by Resnet-50 belong to both the structure as well as the context.

\bibliographystyle{unsrt}  

\clearpage
\bibliography{references}
\bibliographystyle{arxiv}

\end{document}